\documentclass{llncs}
\usepackage{amssymb}
\usepackage{amsmath}
\usepackage{graphicx}
\usepackage{hyperref}
\usepackage{url}

\usepackage{float}
\floatstyle{plaintop}
\restylefloat{table}

\newcommand{\keywords}[1]{\par\addvspace\baselineskip
\noindent\keywordname\enspace\ignorespaces#1}

\usepackage{array}
\newcolumntype{C}[1]{>{\centering\let\newline\\\arraybackslash\hspace{0pt}}m{#1}}
\newcolumntype{L}[1]{>{\let\newline\\\arraybackslash\hspace{0pt}}m{#1}}

\begin{document}

\mainmatter  

\title{Plant identification in an open-world (LifeCLEF 2016)}

\titlerunning{LifeCLEF Plant Identification Task 2016}

\author{Herv\'e Go\"eau\inst{1}
  \and Pierre Bonnet\inst{4}
  \and Alexis Joly\inst{2,3}
}

\tocauthor{Herv\'e Go\"eau, Alexis Joly, Pierre Bonnet}
%
\institute{IRD, UMR AMAP, France,
\email{herve.goeau@cirad.fr}
\and 
Inria ZENITH team, France, 
\email{alexis.joly@inria.fr}
\and
LIRMM, Montpellier, France
\and
CIRAD, UMR AMAP, France,
\email{pierre.bonnet@cirad.fr}
}

\toctitle{LifeCLEF Plant Identification Task 2016}

\maketitle

\begin{abstract}
The LifeCLEF plant identification challenge aims at evaluating plant identification methods and systems at a very large scale, close to the conditions of a real-world biodiversity monitoring scenario. The 2016-th edition was actually conducted on a set of more than 110K images illustrating 1000 plant species living in West Europe, built through a large-scale participatory sensing platform initiated in 2011 and which now involves tens of thousands of contributors. The main novelty over the previous years is that the identification task was evaluated as an \textit{open-set} recognition problem, \textit{i.e.} a  problem in which the recognition system has to be robust to unknown and never seen categories. Beyond the brute-force classification across the known classes of the training set, the big challenge was thus to automatically reject the false positive classification hits that are caused by the unknown classes. This overview presents more precisely the resources and assessments of the challenge, summarizes the approaches and systems employed by the participating research groups, and provides an analysis of the main outcomes.

\end{abstract}

\section{Introduction} 
Image-based plant identification is the most promising solution towards bridging the botanical taxonomic gap, as illustrated by the proliferation of research work on the topic \cite{PlantNetACM2011}, \cite{Tougne2011Leaves}, \cite{kebapci2011plant}, \cite{hazra2013shape}, \cite{aptoula2013morphological} as well as the emergence of dedicated mobile applications such as LeafSnap \cite{leafsnap2} or Pl@ntNet \cite{joly2014interactive}. As promising as these applications are, their performance is still far from the requirements of a fully automated ecological surveillance scenario. Allowing the mass of citizens to produce accurate plant observations requires to equip them with much more effective identification tools. As an illustration, in 2015, 2,328,502 millions queries have been submitted by the users of the Pl@ntNet mobile apps but only less than 3$\%$ of them were finally shared and collaboratively validated. Allowing the exploitation of the unvalidated observations could scale up the world-wide collection of plant records by several orders of magnitude. Measuring and boosting the performance of automated identification tools is therefore crucial. As a first step towards evaluating the feasibility of such an automated biodiversity monitoring paradigm, we created and shared a new testbed entirely composed of image search logs of the Pl@ntNet mobile application (contrary to the previous editions of the PlantCLEF benchmark that were based on explicitly shared and validated plant observations).

As a concrete scenario, we focused on the monitoring of invasive exotic plant species. These species represent today a major economic cost to our society (estimated at nearly 12 billion euros a year in Europe) and one of the main threats to biodiversity conservation \cite{weber2004assessing}. This cost can even be more important at the country level, such as in China where it is evaluated to be about 15 billion US dollars annually \cite{weber2008invasive}, and more than 34 billion US dollars in the US \cite{pimentel2005update}. The early detection of the appearance of these species, as well as the monitoring of changes in their distribution and phenology, are key elements to manage them, and reduce the cost of their management. The analysis of Pl@ntNet search logs can provide a highly valuable response to this problem because the presence of these species is highly correlated with that of humans (and thus to the density of data occurrences produced through the mobile application).

\section{Dataset}

\subsection{Training dataset}
For the training set, we provided the PlantCLEF 2015 dataset enriched with the ground truth annotations of the test images (that were kept secret during the 2015 campaign). More precisely, PlantCLEF 2015 dataset is composed of 113,205 pictures belonging to 41,794 observations of 1000 species of trees, herbs and ferns living in Western European regions. This data was collected by 8,960 distinct contributors. Each picture belongs to one and only one of the 7 types of views reported in the meta-data (entire plant, fruit, leaf, flower, stem, branch, leaf scan) and is associated with a single plant observation identifier allowing to link it with the other pictures of the same individual plant (observed the same day by the same person). An originality of the PlantCLEF dataset is that its social nature makes it close to the conditions of a real-world identification scenario: (i) images of the same species are coming from distinct plants living in distinct areas, (ii) pictures are taken by different users that might not used the same protocol of image acquisition, (iii) pictures are taken at different periods in the year. Each image of the dataset is associated with contextual meta-data (author, date, locality name, plant id) and social data (user ratings on image quality, collaboratively validated taxon name, vernacular name) provided in a structured xml file. The gps geo-localization and device settings are available only for some of the images. More precisely, each image is associated with the followings meta-data:
\begin{itemize}
\item \textbf{ObservationId}: the plant observation ID from which several pictures can be associated
\item \textbf{FileName}
\item \textbf{MediaId}: id of the image
\item View \textbf{Content}: Entire or Branch or Flower or Fruit or Leaf or LeafScan or Stem 
\item \textbf{ClassId}: the class number ID that must be used as ground-truth. It is a numerical taxonomical number used by Tela Botanica
\item \textbf{Species} the species names (containing 3 parts: the Genus name, the specific epithe, the author(s) who discovered or revised the name of the species)
\item \textbf{Genus}: the name of the Genus, one level above the Species in the taxonomical hierarchy used by Tela Botanica
\item \textbf{Family}: the name of the Family, two levels above the Species in the taxonomical hierarchy used by Tela Botanica
\item \textbf{Date}: (if available) the date when the plant was observed
\item \textbf{Vote}: the (round up) average of the user ratings of image quality
\item \textbf{Location}: (if available) locality name, a town most of the time
\item \textbf{Latitude} \& \textbf{Longitude}: (if available) the GPS coordinates of the observation in the EXIF metadata, or if no GPS information were found in the EXIF, the GPS coordinates of the locality where the plant was observed (only for the towns of metropolitan France)
\item \textbf{Author}: name of the author of the picture
\item \textbf{YearInCLEF}: ImageCLEF2011, ImageCLEF2012, ImageCLEF2013, PlantCLEF2014, PlantCLEF2015 specifying when the image was integrated in the challenge
\item \textbf{IndividualPlantId2014}: the plant observation ID used last year during the LifeCLEF2014 plant task
\item \textbf{ImageID2014}: the image id.jpg used in 2014.
\end{itemize}


\subsection{Test dataset}
For the test set, we created a new annotated dataset based on the image queries that were submitted by authenticated users of the Pl@ntNet mobile application in 2015 (unauthenticated queries had to be removed for copyright issues). A fraction of that queries were already associated to a valid species name because they were explicitly shared by their authors and collaboratively revised. We included in the test set the 4633 ones that were associated to a species belonging to the 1000 species of the training set (populating the known classes). Remaining pictures were distributed to a pool of botanists in charge of manually annotating them either with a valid species name or with newly created tags of their choice (and shared between them). In the period of time devoted to this process, they were able to manually annotate 1821 pictures that were included in the test set. Therefore, 144 new tags were created to qualify the unknown classes such as for instance \textit{non-plant objects}, \textit{legs} or \textit{hands}, \textit{UVO} (Unidentified Vegetal Object), \textit{artificial plants}, \textit{cactaceae}, \textit{mushrooms}, \textit{animals}, \textit{food}, \textit{vegetables} or more precise names of horticultural plants such as roses, geraniums, ficus, etc. For privacy reasons, we had to remove all images tagged as \textit{people} (about $1.1\%$ of the tagged queries). Finally, to complete the number of test images belonging to unknown classes, we randomly selected a set of 1546 image queries that were associated to a valid species name that do not belong to the Western European flora (and thus, that do not belong to the 1000 species of the training set or to potentially highly similar species). In the end, the test set was composed of 8,000 pictures, 4633 labeled with one of the 1000 known classes of the training set, and 3367 labeled as new unknown classes. Among the 4633 images of known species, 366 were tagged as \textit{invasive} according to a selected list of 26 potentially invasive species. This list was defined by aggregating several sources (such as the National Botanical conservatory, and the Global Invasive Species Programme) and by computing the intersection with the 1000 species of the training set.

\section{Task Description}
Based on the previously described testbed, we conducted a system-oriented evaluation involving different research groups who downloaded the data and ran their system. To avoid participants tuning their algorithms on the invasive species scenario and keep our evaluation generalizable to other ones, we did not provide the list of species to be detected. Participants only knew that the targeted species were included in a larger set of 1000 species for which we provided the training set. Participants were also aware that (i) most of the test data does not belong to the targeted list of species (ii) a large fraction of them does not belong to the training set of the 1000 species, and (iii) a fraction of them might not even be plants. In essence, the task to be addressed is related to what is sometimes called \textit{open-set} or \textit{open-world} recognition problems \cite{BendaleB14,Scheirer_2014_TPAMIb}, i.e. problems in which the recognition system has to be robust to unknown and never seen categories. Beyond the brute-force classification across the known classes of the training set, a big challenge is thus to automatically reject the false positive classification hits that are caused by the unknown classes (\textit{i.e.} by the distractors). To measure this ability of the evaluated systems, each prediction had to be associated with a confidence score in $p\in[0,1]$ quantifying the probability that this prediction is true (independently from the other predictions).\\
\\
Each participating group was allowed to submit up to 4 runs built from different methods. Semi-supervised, interactive or crowdsourced approaches were allowed but compared independently from fully automatic methods. Any human assistance in the processing of the test queries had therefore to be signaled in the submitted runs.\\
\\
Participants to the challenge were allowed to use external training data at the condition that the experiment is entirely re-producible, i.e. that the used external resource is clearly referenced and accessible to any other research group in the world, and, the additional resource does not contain any of the test observations. It was in particular strictly forbidden to crawl training data from the following domain names:\\
\url{http://ds.plantnet-project.org/}\\
\url{http://www.tela-botanica.org}\\
\url{http://identify.plantnet-project.org}\\
\url{http://publish.plantnet-project.org/}\\
\url{http://www.gbif.org/}\\

\section{Metric}

The metric used to evaluate the performance of the systems is the classification mean Average Precision, called hereinafter "mAP-open", considering each class $c_i$ of the training set as a query. More concretely, for each class $c_i$, we extract from the run file all predictions with $PredictedClassId=c_i$, rank them by decreasing probability $p\in[0,1]$ and compute the Average Precision for that class. The mean is then computed across all classes. Distractors associated to high probability values (i.e. false alarms) are likely to highly degrade the mAP, it is thus crucial to try rejecting them. To evaluate more specifically the targeted usage scenario (\textit{i.e.} invasive species), a secondary mAP ("mAP-open-invasive") was computed by considering as queries only a subset of the species that belong to a black list of invasive species.

\section{Participants and methods}
94 research groups registered to LifeCLEF plant challenge 2016 and downloaded the dataset. Among this large raw audience, 8 research groups succeeded in submitting \textit{runs}, \textit{i.e.} files containing the predictions of the system(s) they ran. Details of the methods and systems used in the runs are further developed in the individual working notes of the participants (Bluefield \cite{Bluefield2016}, Sabanci \cite{Sabanci2016}, CMP \cite{CMP2016}, LIIR, Floristic \cite{Floristic2016}, UM \cite{UM2016}, QUT \cite{QUT2016}, BME \cite{BMETMIT2016}). Table \ref{tab:rawresults} provides the results achieved by each run as well as a brief synthesis of the methods used in each of them. Complementary, the following paragraphs give a few more details about the methods and the overall strategy employed by each participant.\\
\\
\begin{table}
    \centering
    \vspace{3mm}
    \begin{tabular}{|C{24mm}|C{38mm}|C{27mm}|C{10mm}|C{10mm}|C{10mm}|}
    \hline
    Run & Key-words & Rejection & mAP-open & mAP-open-invasive & mAP-closed\\
    \hline
    \hline
Bluefield Run4
& VGGNet, combine outputs from a same observation
& thresholds by class (train+validation)
& 0.742 & 0.717 & 0.827\\
\hline
SabanciU GebzeTU Run1
& 2x(VGGNet,GoogleNet) tuned with resp. 70k, 115k training images
& GoogleNet 70k/70k Plant/ImageNet
& 0.738 & 0.704 & 0.806\\
\hline
SabanciU...Run3
& SabanciUGebzeTU Run1
& Manually removed 90 test images
& 0.737 & 0.703 & 0.807\\
\hline
Bluefield Run3 
& Bluefield Run 4
& thresholds by class
& 0.736 & 0.718 & 0.82\\
\hline
SabanciU...Run2
& SabanciUGebzeTU Run1
& - 
& 0.736 & 0.683 & 0.807\\
\hline
SabanciU...Run4 
& SabanciUGebzeTU Run1
& - 
& 0.735 & 0.695 & 0.802\\
\hline
CMP Run1
& Bagging of 3xResNet-152
& -
& 0.71 & 0.653 & 0.79\\
\hline
LIIR KUL Run3 & CaffeNet, VGGNet16, 3xGoogleNet, adding 12k external plant images & threshold & 0.703 & 0.674 & 0.761\\
\hline
LIIR KUL Run2 
& LIIR KUL Run 3
& threshold
& 0.692 & 0.667 & 0.744\\
\hline
LIIR KUL Run1 
& LIIR KUL Run 3 
& threshold
& 0.669 & 0.652 & 0.708\\
\hline
UM Run4 
& VGGNet16
& -
& 0.669 & 0.598 & 0.742\\
\hline
CMP Run2 
& ResNet-152 
& - 
& 0.644 & 0.564 & 0.729\\
\hline
CMP Run3 
& ResNet-152 (2015training) 
& - 
& 0.639 & 0.59 & 0.723\\
\hline
QUT Run3 
& 1 "general" GoogleNet, 6 "organ" GoogleNets, observation combination 
& -
& 0.629 & 0.61 & 0.696\\
\hline
Floristic Run3 
& GoogleNet, metadata 
& - & 0.627 & 0.533 & 0.693\\
\hline
UM Run1 
& VGGNet16
& -
& 0.627 & 0.537 & 0.7\\
\hline
Floristic Run1 
& GoogleNet 
& - 
& 0.619 & 0.541 & 0.694\\
\hline
Bluefield Run1 
& VGGNet 
& thresholds by class 
& 0.611 & 0.6 & 0.692\\
\hline
Bluefield Run2 
& VGGNet 
& thresholds by class  
& 0.611 & 0.6 & 0.693\\
\hline
Floristic Run2 
& GoogleNet 
& thresholds by class
& 0.611 & 0.538 & 0.681\\
\hline
QUT Run1 
& GoogleNet
& -
& 0.601 & 0.563 & 0.672\\
\hline
UM Run3 
& VGGNet16 with dedicated and combined organ \& species layers
& -  
& 0.589 & 0.509 & 0.652\\
\hline
QUT Run2 
& 6 "organ" GoogleNets, observation combination 
& 
& 0.564 & 0.562 & 0.641\\
\hline
UM Run2 
& VGGNet16 from scratch (without ImageNet2012) 
& -
& 0.481 & 0.446 & 0.552\\
\hline
QUT Run4 
& QUT Run3
& threshold 
& 0.367 & 0.359 & 0.378\\
\hline
BMETMITRun4 
& AlexNet \& BVWs \& metadata
& - 
& 0.174 & 0.144 & 0.213\\
\hline
BMETMITRun3 
& AlexNet \& BVWs  \& metadata
& threshold by classifier
& 0.17 & 0.125 & 0.197\\
\hline
BMETMITRun1 
& AlexNet 
& - 
& 0.169 & 0.125 & 0.196\\
\hline
BMETMITRun2 
& BVWs (fisher vectors)
& - 
& 0.066 & 0.128 & 0.101\\
\hline
\end{tabular}
\caption{Results of the LifeCLEF 2016 Plant Identification Task. Column "Key-words" \& "Rejection" attempt to give the main idea of the method used.}
\label{tab:rawresults}
\end{table}
\textbf{Bluefield system, Japan, 4 runs, \cite{Bluefield2016}}: A VGGNet \cite{Simonyan14c} based system with the addition of Spatial Pyramid Pooling, Parametric ReLU and unknown class rejection based on the minimal prediction score of training data (Run 1). Run 2 is the same as run 1 but with a slightly different rejection making use of a validation set. Run 3 and 4 are respectively the same as Run 1 and 2 but the scores of the images belonging to the same observation were summed and normalised.\\
\\
\textbf{BME TMIT system, Hungary, 4 runs, \cite{BMETMIT2016}}: This team attempted to combine three classification methods: (i) one based on dense SIFT features, fisher vectors and SVM (Run 2), (ii) the second one based on AlexNet CNN (Run 1) and (iii), the last one based on a SVM trained on the meta-data. Run 3 corresponds to the combination of three classifiers (using a weighted average) and Run 4 added two rejection mechanisms to Run3 (a distance-based rejection for the fisher vectors and the minimal prediction score of training data for the CNN).\\
\\
\textbf{CMP system, Czech Republic, 3 runs}: This team built their system with the very deep residual CNN approach ResNet with 152 layers \cite{he2015deep} which achieved the best results in both ILSVRC 2015 and COCO 2015 (Common Objects in Context) challenges last year. They added a fully-connected layer with 512 neurons on top of the network, right before softmax classifier with an maxout activation function \cite{Maxoutnetworks}. They obtained thus a first run (run 2) by using all the 2016 training dataset while they used only the 2015 training dataset in run 3. Run 1 achieved the best performances by using a bagging approach of 3 ResNet-152: the training dataset was divided into three folds, and each CNN was using a different fold for validation and the remaining two folds for fine tuning.\\
\\
\textbf{Floristic system, France, 3 runs, \cite{Floristic2016}}: This participant used a modified GoogleNet architecture by adding batch normalisation and ReLU activation function instead of the PReLU ones (run 1). In Run 2, adaptive thresholds (one for each class) based on the prediction of the training images in the fine tuned CNN were estimated for removing too low prediction on test images. Run 3 used a visual similarity search for scaling down the initial CNN prediction when a test image gives inhomogeneous knns according to the metadata (organ tags, GPS, genus and family levels).\\
\\
\textbf{LIIR KUL system, Belgium, 3 runs}: This team used a ensemble classifier of 5 fine-tuned models: one CaffeNet, one VGGNet16 and 3 GoogLeNet. They added 12k external training data from Oxford flowers set, LeafSnap and trunk12 and attempted to exploit information in the metadata, mostly range maps from GPS coordinates comparing predictions with content tags. As a rejection criteria, they used a threshold on confidence of best prediction, one different threshold for each run (run 1: 0.25, run 2: 0.2, run 3: 0.15).\\
\\
\textbf{QUT system, Australia, 4 runs, \cite{QUT2016}}: This participant compared a standard CNN fine tuned approach based on GoogleNet (run 1) with a bagging approach "mixDCNN" (run 2) built on the top of 6 fine tuned GoogleNet on the 6 training subsets corresponding to the 6 distinct organs ("leaf" and "leafscan" training images are actually merged into one subset). Outputs are weighted by "occupation probabilities" which give for each CNN a confidence about their prediction. Run 3 merged the two approaches, run 4 too but with a threshold attempting to remove false positives.\\
\\
\textbf{Sabanci system, Turkey, 4 runs, \cite{Sabanci2016}}: This team used a CNN-based system with 2 main configurations. Run 1: an ensemble of GoogleLeNet \cite{szegedy2015going} and VGGNet \cite{Simonyan14c} fine-tuned on LifeCLEF 2015 data (for recognizing the targeted species), as well as a second GoogleNet fine-tuned on the binary rejection problem (using 70k images of PlantCLEF2016 training set for the known class label and 70K external images from the ILSCVR dataset for the unknown class label). Run 2 is the same than Run 1 but without rejection. Run 3 is the same than Run 1 but with manual rejection of 90 obviously non plant images.\\
\\
\textbf{UM system, Malaysia \& UK, 4 runs}: This team used a CNN system based on a VGGNet 16 layers. They modified the higher convolutional level in order to learn at the same time combinations of species and organs. VGGNet16 with dedicated and combined organ \& species layers with seven organ labels: branch, entire, flower, fruit, leaf, leafscan and stem.
\\
\\

\section{Official Results}
\begin{figure}[!t]
\centering
\includegraphics[width=0.95\linewidth]{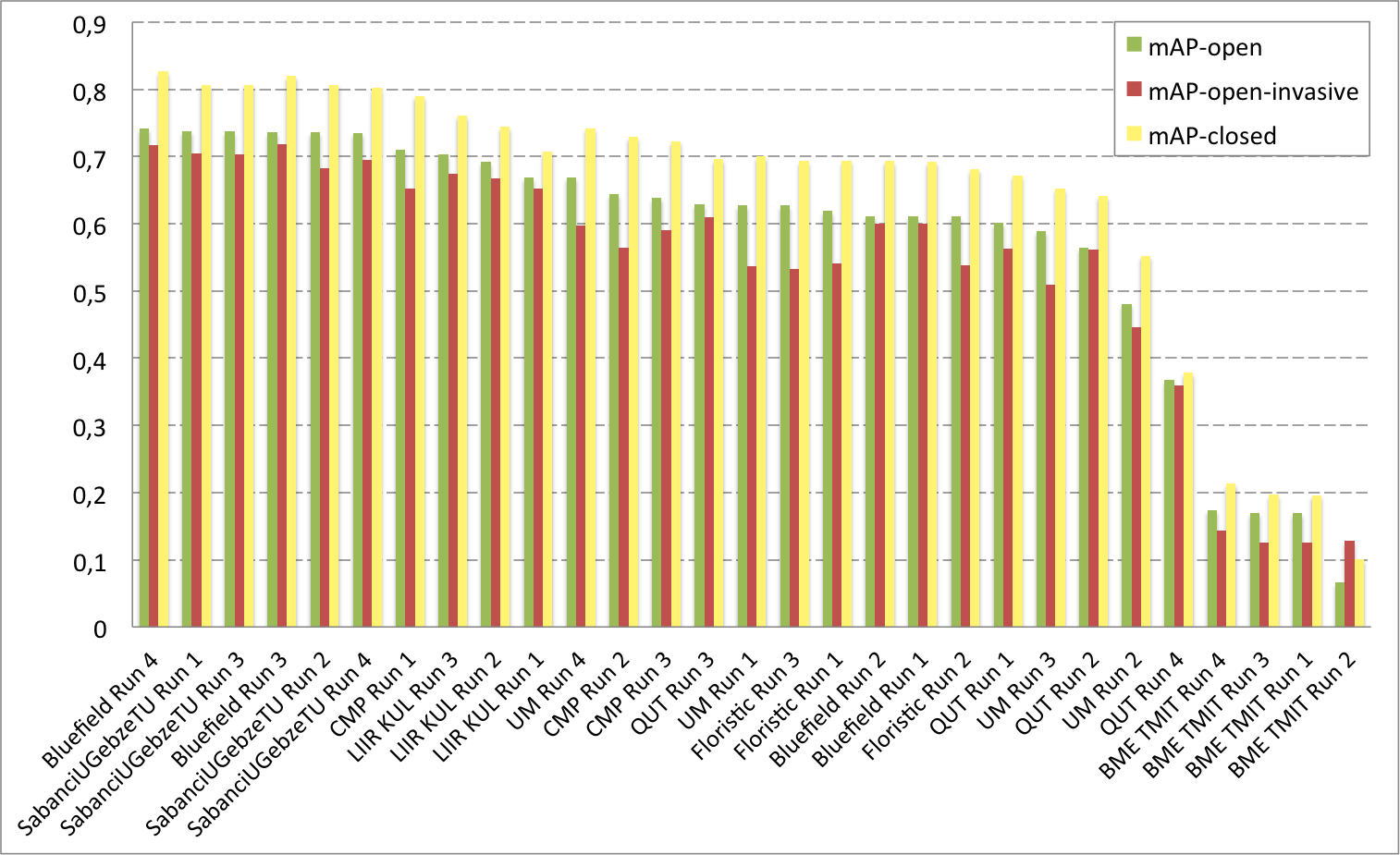}
\caption{Scores achieved by all systems evaluated within the plant identification task of LifeCLEF 2016, \textbf{mAP-open}: mean Average Precision on the 1000 species of the training set and distractors in the test set, \textbf{mAP-open-invasive}: mean Average Precision with distractors but restricted to 26 invasive species only, \textbf{mAP-closed}: mean Average Precision on the 1000 species but without distractors in the test set}
\label{fig:PlantCLEF2016OfficialScoreCMAP}
\end{figure}

We report in Figure \ref{fig:PlantCLEF2016OfficialScoreCMAP} the scores achieved by the 29 collected runs for the two official evaluation metrics (mAP-open and mAP-open-invasive). To better assess the impact of the distractors (\textit{i.e.} the images in the test set belonging to unknown classes), we also report the mAP obtained when removing them (and denoted as mAP-closed). As a first noticeable remark, the top-26 runs which performed the best were based on Convolutional Neural Networks (CNN). This definitely confirms the supremacy of deep learning approaches over previous methods, in particular the one bases on hand-crafted features (such as BME TMIT Run 2). The different CNN-based systems mainly differed in (i) the architecture of the used CNN, (ii) the way in which the rejection of the unknown classes was managed and (iii), various system design improvements such as classifier ensembles, bagging or observation-level pooling. An impressive mAP of $0.718$ (for the targeted invasive species monitoring scenario) was achieved by the best system configuration of Bluefield (run 3). The gain achieved by this run is however more related to the use of the observation-level pooling (looking at Bluefield run 1 for comparison) than to a good rejection of the distractors. Comparing the metric mAP-open with mAP-closed, the figure actually shows that the presence of the unknown classes degrades the performance of all systems in a roughly similar way. This difficulty of rejecting the unknown classes is confirmed by the very low difference between the runs of the participants who experimented their system with or without rejection (\textit{e.g.} Sabanci Run 1 vs. Run 2 or FlorisTic Run 1 vs. Run 2). On the other side, one can remark that all systems are quite robust to the presence of unknown classes since the drop in performance is not too high. Actually, as all the used CNNs were pre-trained on a large generalist data set beforehand (ImageNet), it is likely that they have learned a diverse enough set of visual patterns to avoid underfiting.

\section{Complementary Analysis: Impact of the degree of novelty}
Within the conducted evaluation, the proportion of unknown classes in the test set was still reasonable (actually only 42$\%$) because of the procedure used to create it. In a real mobile search data stream, the proportion of images belonging to unknown classes could actually be much higher. To simulate such a higher degree of novelty, we progressively down sampled the test images belonging to known classes and recomputed the mAP-open evaluation metric. Results of this experiment are provided in Figure. For clarity, we only reported the curves of the best systems (for various degrees of novelty). As a first conclusion, the chart clearly shows that the degree of novelty in the test set has a strong influence on the performance of all systems. Even when $25\%$ of the queries still belong to a known class, none  of the evaluated systems reach a mean average precision greater than $0.45$ (to be compared to $0.83$ in a closed world). Some systems do however better resist to the novelty than others. The performance of the best run of Bluefield on the official test set does for instance quickly degrade with higher novelty rates (despite the use of a rejection strategy). Looking at the best run of Sabanci, one can see that the use of a supervised rejection class is the most beneficial strategy for moderate novelty rates but then the performance also degrades for high rates. Interestingly, the comparison of LIIR KUL Run1 and LIIR KUL Run3 show that simply using a higher rejection threshold applied to the CNN probabilities is more beneficial in the context of high unknown class rates. Thus, we believe there is still rooms of improvements in the design of adaptive rejection methods that would allow to automatically adapt the strength of the rejection to the degree of novelty.
 
\begin{figure}[!t]
\centering
\includegraphics[width=0.95\linewidth]{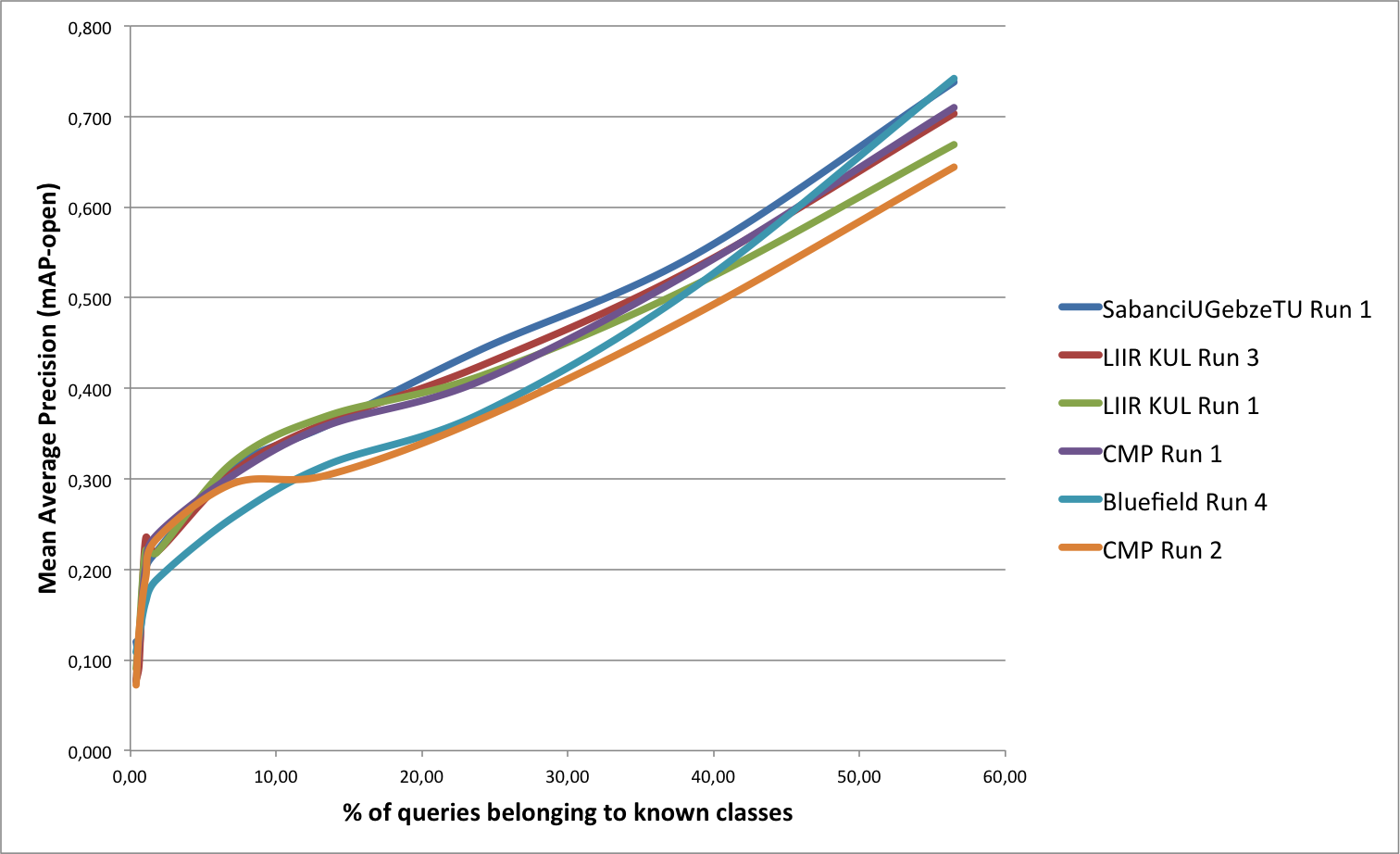}
\vspace{-10pt}
\caption{Impact of the degree of novelty: Mean Average Precision vs. proportion of test images belonging to known classes}
\label{fig:precision-recall-best}
\end{figure}

\section{Conclusion}
This paper presented the overview and the results of the LifeCLEF 2016 plant identification challenge following the five previous ones conducted within CLEF evaluation forum. The main novelty compared to the previous year was that the identification task was evaluated as an \textit{open-set} recognition problem, \textit{i.e.} a  problem in which the recognition system has to be robust to unknown and never seen categories. The main conclusion was that CNNs appeared to be naturally quite robust to the presence of unknown classes in the test set but that none of the rejection methods additionally employed by the participants improved that robustness. Also, the proportion of novelty in the test was still moderate. We therefore conducted additional experiments showing that the preformance of CNNs is strongly affected by higher rates of images belonging to unknown classes and that the problem is clearly still open. In the end, our study shows that there is still some room of improvement before being able to share automatically identified plant observations within international biodiversity platforms. The proportion of false positives would actually be too high for being acceptable for biologists.
\bibliographystyle{splncs03}

\end{document}